%% file: main.tex
\documentclass[conference]{IEEEtran}
\usepackage{times}
\usepackage[numbers]{natbib}
\usepackage{multicol}
\usepackage[bookmarks=true]{hyperref}
\usepackage{placeins}
\usepackage{amsmath,amssymb,amsfonts}
\usepackage{graphicx}
\usepackage{booktabs}
\usepackage{multirow}
\usepackage{array}
\usepackage{colortbl}
\usepackage[table]{xcolor}
\usepackage{arydshln}
\usepackage{subcaption}
\usepackage{caption}
\usepackage{wrapfig}
\usepackage{float}
\usepackage{placeins}
\usepackage{pifont}
\usepackage{url}
\usepackage{fontawesome5}

\providecommand{\vecsym}[1]{\mathbf{#1}}
\definecolor{bestcell}{RGB}{255,242,204}
\newcommand{\best}[1]{\cellcolor{bestcell}\textbf{#1}}

\title{TFP: Temporally Conditioned Memory-Fusion Policies for Visuomotor Learning}

\author{
\IEEEauthorblockN{
Yushen Liang\textsuperscript{1,\ensuremath{\dagger}},
Yue Peng\textsuperscript{1,\ensuremath{\dagger}},
Baosheng Jin\textsuperscript{1,\ensuremath{\dagger}},
Tianluo Zhang\textsuperscript{1},\\
Xinyu Zhang\textsuperscript{2},
Shuyi Zhou,
Zhuoran Chen\textsuperscript{1},
Xinqi Liu\textsuperscript{1},
Shenji Wan\textsuperscript{1}
}

\IEEEauthorblockA{\textsuperscript{1}NYU Shanghai, Shanghai, China}
\IEEEauthorblockA{\textsuperscript{2}University of Electronic Science and Technology of China, Chengdu, China}

\IEEEauthorblockA{\textsuperscript{\ensuremath{\dagger}}Equal contribution}
\IEEEauthorblockA{%
\small
\href{https://github.com/Mirage415/TFP_pro}
{\faIcon{github}\ \textbf{Code:} \nolinkurl{github.com/Mirage415/TFP-Temporally-conditioned-Memory-Fusion-Policies-for-Visuomotor-Learning}}%
}
}

\begin{document}
\maketitle

\input{sections/Abstract.tex}
\IEEEpeerreviewmaketitle
\input{sections/TeaserFigure.tex}

\input{sections/Introduction.tex}
\input{sections/RelatedWork.tex}
\input{sections/Method.tex}
\input{sections/Experiments.tex}
\input{sections/Conclusion.tex}

\clearpage
\input{sections/References.tex}

\clearpage
\input{appendices/Appendix.tex}

\end{document}

%% file: sections/Abstract.tex
\begin{abstract}
Vision--Language--Action (VLA) policies such as $\pi_{0.5}$ and OpenVLA perform well on many manipulation tasks, but they are often reactive: the next action is predicted from the current observation, instruction, and proprioceptive state. This assumption breaks down in stage-dependent manipulation, where visually similar states may require different actions depending on latent task progress and previous interaction outcomes. We argue that such tasks require not only memory, but dynamics-aware belief updates: the policy should preserve task progress during stable or occluded phases and revise its belief near contact, release, or subgoal transitions.  We introduce Temporally Conditioned Memory-Fusion Policies (TFP), a lightweight memory-action framework for VLA backbones. TFP maintains an episode-local task-progress belief with Liquid Time-Constant dynamics and injects the updated belief directly into the flow-matching action decoder through adaptive modulation. This lets temporally accumulated context shape the generated action chunk, rather than serving only as passive history context. With a 3.3B-parameter model, TFP improves the average success rate from \(96.9\%\) to \(98.75\%\) on LIBERO and from \(91.4\%\) to \(93.77\%\) on LIBERO-plus. On the memory-focused MIKASA ShellGameTouch diagnostic, TFP achieves success up to \(75.0\%\). Mechanistic analyses show that write-gain changes near manipulation events are about \(6\times\) larger than far non-event phases, and hidden-state interventions show that the belief causally modulates generated action chunks. These results suggest that compact, event-sensitive memory dynamics can improve VLA policies under occlusion, visual perturbation, and stage-dependent task structure.
\end{abstract}


%% file: sections/TeaserFigure.tex
\begin{figure*}[t]
    \centering
    \includegraphics[width=\textwidth]{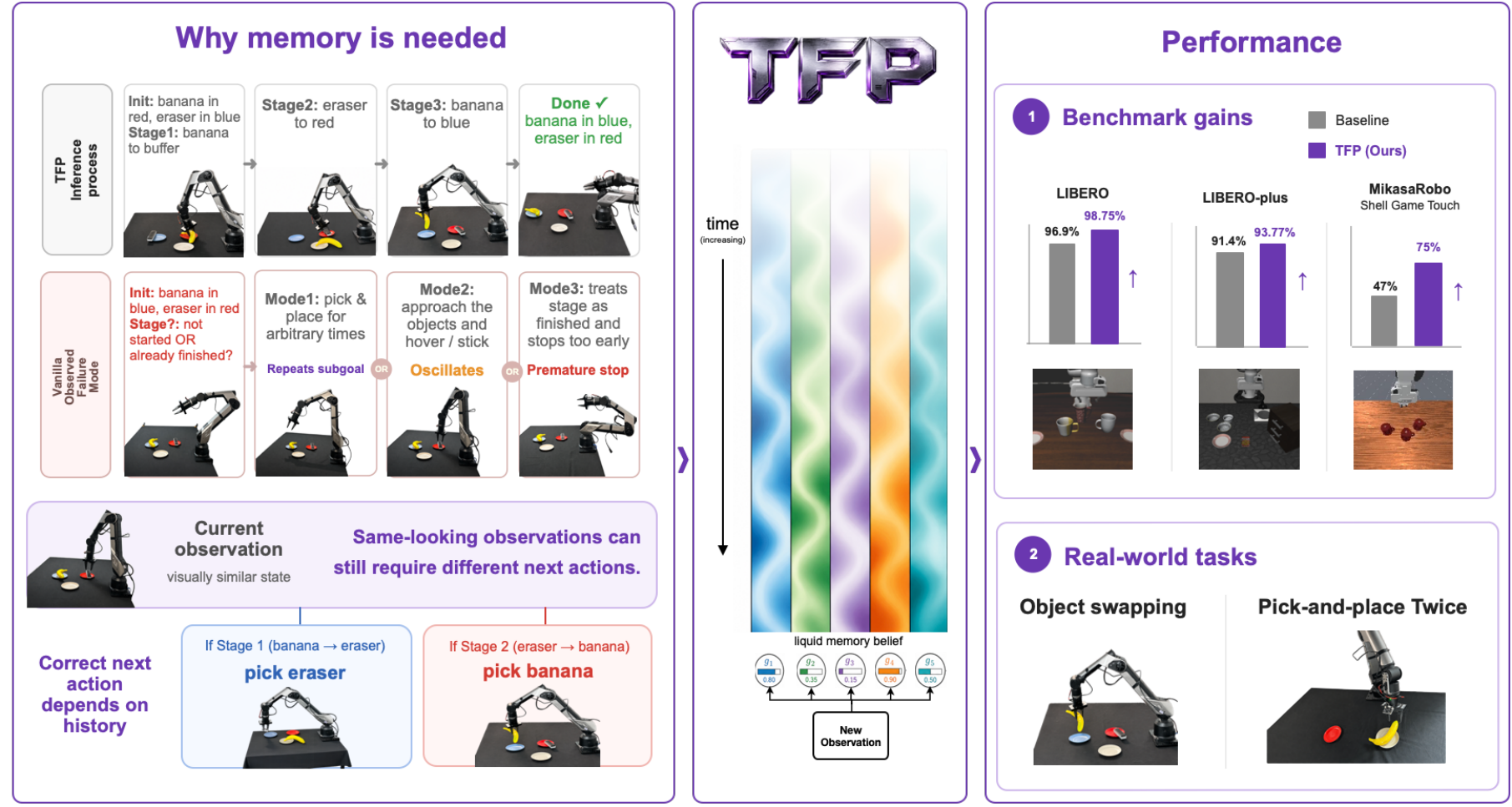}
    \caption{Overview illustrating the motivation for memory-conditioned visuomotor control.}
    \label{fig:intro_overview}
\end{figure*}

%% file: sections/Introduction.tex
\section{Introduction}

Vision--Language--Action (VLA) models, such as $\pi_{0.5}$, OpenVLA,
and Octo, have achieved strong performance in robotic manipulation by mapping
language instructions and multimodal observations directly to actions
~\cite{black2024pi0,brohan2023rt2,driess2023palme,openvla2024,octo2024}.
However, many VLA policies remain largely reactive: each action is predicted
from the current observation, instruction, and proprioceptive state. This is
insufficient for stage-dependent manipulation, where the same visible scene may
require different actions depending on latent task progress and previous
interaction outcomes. For example, in object swapping, a visually similar state
may correspond to moving the first object to a buffer, moving the second object
into the first object's original location, or terminating the task. 

This motivates adding memory, or task belief, to VLA policies. We use
``belief'' to denote a learned latent memory state that summarizes hidden task
progress relevant to future actions. However, robotic manipulation requires more than storing past observations. A useful belief must also decide \emph{when} to change. During
stable transport or occlusion, the policy should preserve task-progress
information despite ambiguous visual evidence; near contact, release, or
subgoal completion, it should rapidly incorporate new evidence. Thus, the key
question is not only whether a VLA has memory, but whether its memory update
follows the event structure and temporal dynamics of manipulation.

Existing memory-aware robot policies mainly either retrieve from history buffers
or maintain recurrent latent states~\cite{shi2025memoryvla,prism2026,mem2026,
xiao2025avavla,remem_vla2026}. History-based methods preserve rich episodic
detail but do not explicitly form a compact task-progress state; recurrent
methods provide such a state, but updates are typically indexed by policy updates rather than manipulation events and elapsed
physical time.

We argue that chunked visuomotor control calls for a dynamics-aware belief
update. A practical controller may re-query earlier when a chunk becomes unreliable, unstable, or
requires correction after contact. Thus, the interval between policy queries is
a physical-time variable, not merely a discrete step index. 
Liquid Time-Constant (LTC) networks provide a natural mechanism for such belief
tracking. Their hidden states evolve through input-dependent time constants,
allowing the model to retain information when task context should persist and to
update more rapidly when new evidence appears~\cite{hasani2021ltc,hasani2022cfc}.
This continuous-time inductive bias is well matched to memory-dependent
manipulation, where completed subgoals, prior contacts, and interaction outcomes
may remain relevant across multiple action chunks.

We introduce \textbf{Temporally Conditioned Memory-Fusion Policies (TFP)}, a
belief-conditioned VLA framework for robotic manipulation. TFP maintains an
episode-local latent belief with LTC dynamics: the previous hidden state carries
retained task context, the current visual and proprioceptive observation proposes
an update, and elapsed time controls how strongly the belief is retained or
revised. The updated belief is projected into the action-head conditioning space
and injected through AdaLN-style modulation, allowing temporal context to
directly shape the flow-matching action distribution rather than serve only as
passive historical context.

We evaluate TFP on simulation benchmarks and real-world memory-dependent
manipulation tasks. TFP consistently improves over reactive, history-based, and
latent-state baselines. Mechanistic analyses further show that the learned
belief stabilizes task-progress representations and causally modulates generated
action chunks. Our contributions are:
\begin{itemize}
    \setlength{\itemsep}{0pt}
    \setlength{\parsep}{0pt}
    \setlength{\topsep}{0.15em}
    \setlength{\partopsep}{0pt}
    \item We formulate stage-dependent manipulation as dynamics-aware belief
tracking, where the correct action depends not only on latent task progress and
interaction history, but also on when the internal belief should be revised or
preserved.
    \item We propose TFP, a belief-conditioned VLA framework that maintains a
    continuous-time latent belief using LTC dynamics and injects this belief
    directly into flow-matching action generation. We further introduce
    Episode-Aware Temporal Batching (EATB) to train recurrent belief states
    efficiently while preserving episode-local hidden-state continuity.
    \item We show that TFP improves memory-dependent control in simulation and
    on a real Galaxea A1 robot, and provide mechanistic analyses showing that
    the learned belief stabilizes task-progress representations and causally
    modulates generated action chunks.
\end{itemize}

%% file: sections/RelatedWork.tex
\section{Related Work}
\label{sec:related_work}

\paragraph{Memory in vision-language-action policies.}
Recent VLA policies such as RT-2, OpenVLA, Octo, and $\pi_0$ map language,
visual observations, and robot states directly to actions
~\cite{brohan2023rt2,openvla2024,octo2024,black2024pi0}. Since many
manipulation tasks are partially observable, several works augment such
policies with history or memory. One class stores past observations, actions,
or semantic summaries and retrieves or attends to them at decision time, as in
HAMLET, MemoryVLA, and Causal Diffusion Policy
~\cite{koo2025hamlet,shi2025memoryvla,cdp2025}. These methods improve access to
past context, but memory is mainly used as a retrieval source rather than as an
explicit state of task progress. A second class maintains a compact latent state. AVA-VLA uses a recurrent
belief state to modulate visual processing, and ReMem-VLA propagates recurrent
memory queries across frames and action chunks~\cite{xiao2025avavla,remem_vla2026}.
These methods provide explicit memory, but their updates are typically tied to
frames, policy steps, or chunks. In chunked receding-horizon control, however,
the interval between policy queries can vary with execution progress, contact,
or instability. The memory update should therefore depend on physical elapsed
time, not only on a discrete step index.

\paragraph{Continuous-time belief for chunked control.}
Liquid Time-Constant networks and closed-form continuous-time networks model
hidden-state evolution with input-dependent time constants
~\cite{hasani2021ltc,hasani2022cfc}. TFP uses this continuous-time mechanism as
a task-belief filter for chunked VLA control. The belief retains slow
task-progress information, updates when new visual and proprioceptive evidence
arrives, and directly modulates the flow-matching action decoder through
AdaLN-style conditioning. Thus, memory in TFP is not only retrieved context or an
auxiliary perception signal, but a compact physical-time belief state that
directly shapes action generation. More discussion on broader related work can be found in Appendix~\ref{app:related_work}.

%% file: sections/Method.tex
\section{Method}
\label{sec:method}

\providecommand{\vecsym}[1]{\mathbf{#1}}

TFP augments a chunked VLA policy with an episode-local latent belief over task
progress and interaction history. At each policy query, the current visual and
proprioceptive observations update this belief through Liquid Time-Constant
(LTC) dynamics, and the updated belief directly conditions the flow-matching
action head through AdaLN-style modulation. This couples memory and action
generation: temporal context is not stored as a passive history buffer, but as a
persistent state that modulates the next generated action chunk.

\begin{figure*}[t]
    \centering
    \includegraphics[width=\textwidth]{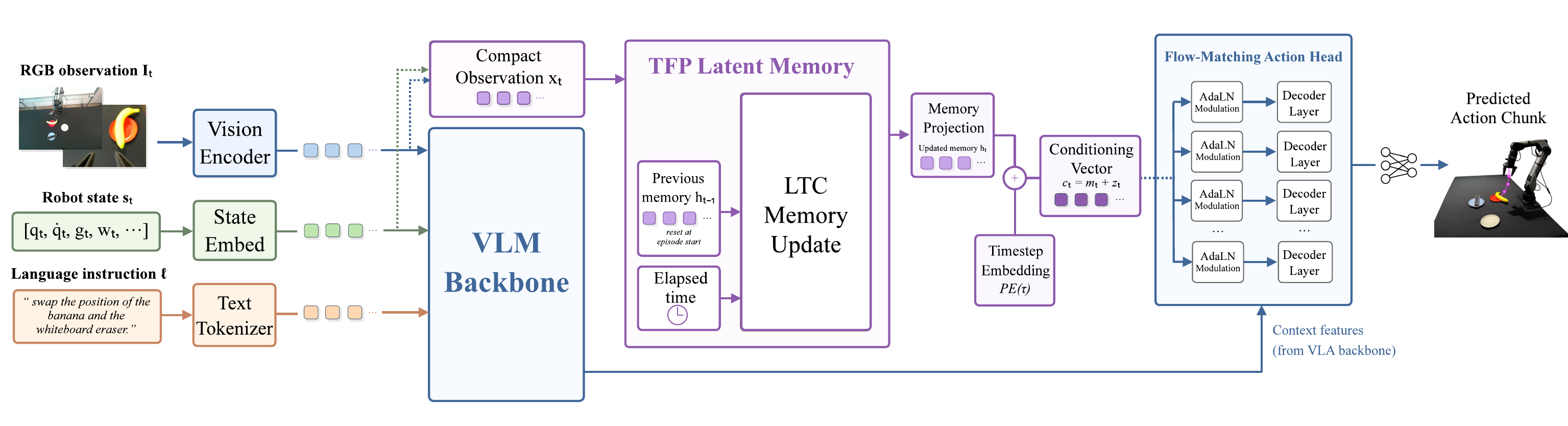}
    \caption{TFP maintains a continuous-time latent belief and injects it into the action decoder through adaptive modulation.}
    \label{fig:architecture}
\end{figure*}

\subsection{Problem Setting and Notation}

At policy query $t$, the agent receives a visual observation $\mathcal{I}_t$,
robot state $\vecsym{s}_t \in \mathbb{R}^{d_s}$, elapsed time
$\Delta t_t \in \mathbb{R}_{\ge 0}$ since the previous query, language
instruction $\ell$, and previous belief $\vecsym{h}_{t-1}\in\mathbb{R}^{d_h}$.
A vision encoder maps $\mathcal{I}_t$ to visual tokens
\begin{equation}
V_t=[\vecsym{v}_{t,1},\dots,\vecsym{v}_{t,N}]
\in\mathbb{R}^{N\times d_v}.
\end{equation}
The policy updates the latent belief $\vecsym{h}_t$ and predicts an action chunk
conditioned on the current observation, robot state, instruction, and belief.

\subsection{Continuous-Time Belief Update}

We first form a compact observation representation
$\vecsym{x}_t=[\phi_{\mathrm{vision}}(V_t);\,\phi_{\mathrm{state}}(\vecsym{s}_t)]$,
where $\phi_{\mathrm{vision}}(\cdot)$ pools or projects visual tokens and
$\phi_{\mathrm{state}}(\cdot)$ embeds the robot state. TFP updates the latent
belief with an LTC-style recurrence:
\begin{equation}
\begin{aligned}
\hat{\vecsym{h}}_t
&=
\tanh\!\left(W_h[\vecsym{x}_t;\vecsym{h}_{t-1}] + b_h\right),
\\
\vecsym{\tau}_t
&=
\operatorname{softplus}\!\left(
W_\tau[\vecsym{x}_t;\vecsym{h}_{t-1}] + b_\tau
\right)+\epsilon .
\end{aligned}
\label{eq:ltc_pre}
\end{equation}

\begin{equation}
\begin{aligned}
\vecsym{k}_t
&=
\exp\!\left(-\Delta t_t/\vecsym{\tau}_t\right),
\\
\vecsym{h}_t
&=
\vecsym{k}_t\odot\vecsym{h}_{t-1}
+
(1-\vecsym{k}_t)\odot\hat{\vecsym{h}}_t .
\end{aligned}
\label{eq:ltc_update}
\end{equation}
Here $\hat{\vecsym{h}}_t$ is the candidate belief induced by the current
observation and previous belief, $\vecsym{\tau}_t\in\mathbb{R}^{d_h}_{>0}$ is a
vector of input-dependent time constants, and $\vecsym{k}_t$ controls how much
of the previous belief is retained over the elapsed interval $\Delta t_t$.
Equivalently, with write gain $\vecsym{g}_t=1-\vecsym{k}_t$, the update can be
written as
\begin{equation}
\vecsym{h}_t-\vecsym{h}_{t-1}
=\vecsym{g}_t\odot(\hat{\vecsym{h}}_t-\vecsym{h}_{t-1}).
\label{eq:actual_update}
\end{equation}
Thus, memory changes only when the candidate belief differs from the retained
belief and the elapsed-time-dependent write gain permits the update. This is
useful for manipulation: slow channels can preserve task context across action
chunks, while faster channels can absorb new perceptual evidence when the scene
changes.
We use \emph{belief} in a functional sense: $\vecsym{h}_t$ is a learned
latent state summarizing task progress and interaction history, rather than an
explicit Bayesian posterior or a raw buffer of past observations.

This shows that memory changes only when the current candidate belief differs
from the retained belief and the write gain allows new evidence to be
incorporated. In manipulation, this is useful because observations can be noisy,
partially occluded, or locally ambiguous. Once the policy has formed a
task-conditioned belief, later observations may produce candidate beliefs that
are already consistent with the retained state, yielding small updates without
making the memory uninformative.

The vector-valued time constant $\tau_t$ allows different latent channels to
operate at different effective time scales. We do not assume that individual
coordinates correspond to human-defined events; rather, task progress, object
state, contact evidence, and occlusion-related information may be represented as
distributed latent patterns. Some subspaces can therefore behave as
high-retention channels for slow task context, while others can update more
quickly with local perceptual evidence.

For each latent dimension $j$, the update unrolls as
\begin{equation}
h_{t,j}
=
\left(\prod_{r=1}^{t} k_{r,j}\right)h_{0,j}
+
\sum_{i=1}^{t}
(1-k_{i,j})
\left(\prod_{r=i+1}^{t} k_{r,j}\right)
\hat h_{i,j}.
\end{equation}
Thus, the memory forms a path-dependent mixture of past candidate beliefs:
a candidate contributes only if it is written when observed and retained by
later updates. This explains the write-then-stabilize behavior observed in our
diagnostics. A useful memory trajectory need not keep changing throughout a
rollout; after informative observations establish the task belief, high
retention can preserve it through occlusion, contact noise, or visually
ambiguous stages.

\subsection{Belief-Conditioned Action Generation}

TFP injects the updated belief into the flow-matching action decoder through
adaptive normalization rather than explicit memory-token cross-attention. Given
$\vecsym{h}_t$, we project it into the decoder conditioning space:
\begin{equation}
\vecsym{m}_t = W_m\vecsym{h}_t+\vecsym{b}_m \in \mathbb{R}^{d_c}.
\end{equation}
The flow-matching timestep is embedded as
\begin{equation}
\vecsym{e}_\tau=\operatorname{PE}(\tau),\qquad
\vecsym{z}_\tau =
W_2\sigma(W_1\vecsym{e}_\tau+\vecsym{b}_1)+\vecsym{b}_2,
\end{equation}
where $\sigma(\cdot)$ is a swish nonlinearity. We combine timestep and belief
conditioning by
\begin{equation}
\vecsym{c}_t=\vecsym{z}_\tau+\vecsym{m}_t,
\qquad
\hat{\vecsym{x}}^{(\ell)}
=
\operatorname{AdaLN}(\vecsym{x}^{(\ell)},\vecsym{c}_t),
\label{eq:adaln_cond}
\end{equation}
where $\vecsym{x}^{(\ell)}$ is the hidden state of decoder layer $\ell$.
AdaLN predicts feature-wise affine modulation parameters from $\vecsym{c}_t$
and applies them to normalized decoder activations
~\cite{Peebles_2023_ICCV,perez2018film}. The resulting action head predicts
the memory-conditioned flow velocity
\begin{equation}
v_\theta(a_\tau,\tau\mid \mathcal{I}_t,\vecsym{s}_t,\ell,\vecsym{h}_t).
\end{equation}
This interface makes the belief directly modulate the generated action chunk,
instead of requiring the decoder to retrieve and interpret a separate memory
token sequence. Consequently, visually similar observations can produce
different actions when their retained task beliefs differ.
Additional discussion of this adaptive-conditioning interface is given in
Appendix~\ref{app:memory_conditioning_interface}.

\subsection{Training with Episode-Aware Temporal Batching}

Training TFP requires preserving hidden-state continuity within each
demonstration episode. Randomly shuffling individual chunks breaks this
continuity, while full-episode backpropagation is memory-intensive for large VLA
backbones. We therefore introduce \emph{Episode-Aware Temporal Batching}
(EATB), which trains on temporally contiguous chunks from multiple episodes in
parallel while maintaining a separate hidden state for each active episode.

For each episode $e$, EATB keeps an episode-local hidden state
$\vecsym{h}_t^{(e)}=f_\theta(\vecsym{h}_{t-1}^{(e)},o_t^{(e)},\Delta t_t^{(e)})$
and resets it only at episode boundaries. At each training step, we sample
$B$ active episodes and unroll each for $K$ consecutive chunks. If episode
$e_b$ starts from chunk index $\tau_b$, its stored hidden state is retrieved as
$\vecsym{h}_{b,0}=\tilde{\vecsym{h}}_{\tau_b}^{(e_b)}$ and updated through the
segment. We optimize the average flow-matching imitation loss over the resulting
$BK$ chunks,
\begin{equation}
\mathcal{L}_s
=
\frac{1}{BK}
\sum_{b=1}^{B}
\sum_{k=1}^{K}
\ell_{\mathrm{flow}}^{(b,k)}.
\end{equation}
After the parameter update, the final hidden state is written back with gradient
truncation:
\begin{equation}
\tilde{\vecsym{h}}_{\tau_b+K}^{(e_b)}
\leftarrow
\operatorname{stopgrad}(\vecsym{h}_{b,K}).
\end{equation}
Thus, hidden states persist numerically across training segments, while
gradients are truncated at segment boundaries. EATB preserves long-horizon
forward memory without the cost of full-episode backpropagation.
Further implementation details are described in
Appendix~\ref{app:method_details}.

\subsection{Adaptive Receding-Horizon Execution}
\label{app:adaptive_executor}

To evaluate TFP under irregular policy-query intervals, we use an inference-time
adaptive receding-horizon executor. The policy always predicts an action chunk
of horizon $H$, but the executor selects a prefix length
$E_t\in[E_{\min},H]$ to execute before the next policy query. The next memory
update receives the actual elapsed interval
\begin{equation}
\Delta t_{t+1}=E_t\delta t_{\mathrm{ctrl}} .
\end{equation}
The executor executes the full horizon by default and shortens the prefix only
when the predicted chunk contains a gripper-transition boundary or a sustained
high-risk region. The per-step chunk risk is
\begin{equation}
R_{t,r}
=
\lambda_j J_{t,r}
+
\lambda_b B_{t,r}
+
\lambda_c C_{t,r},
\end{equation}
where $J_{t,r}$ measures action jerk, $B_{t,r}$ indicates gripper-transition
boundaries, and $C_{t,r}$ measures continuity with the previously executed
action. This changes only the inference-time execution schedule; the policy
network, predicted action horizon, and model weights are unchanged.

\section{Comparison to Recurrent and State-Space Baselines}
\label{app:baseline_comparison}

\subsection{Relation to GRUs}
\label{sec:gru_comparison}

A standard GRU updates its hidden state as
\begin{equation}
h_t
=
z_t \odot h_{t-1}
+
(1-z_t)\odot \tilde{h}_t,
\qquad
z_t
=
\sigma\!\left(W_z[x_t,h_{t-1}] + b_z\right).
\label{eq:gru_update}
\end{equation}
This update resembles the interpolation form used by TFP, but its retention
gate is indexed by recurrent steps rather than physical time. In particular,
when elapsed time is not provided as an input, two updates with the same
observation and previous hidden state necessarily produce the same gate,
even if they occur after different physical intervals.

This differs from continuous-time belief dynamics. For example, in the
absence of new evidence, exponential decay evolves as
\begin{equation}
h_t
=
\exp(-\lambda \Delta t_t)h_{t-1},
\qquad
\lambda > 0,
\label{eq:pure_decay}
\end{equation}
and therefore produces different retained states for different values of
$\Delta t_t$. A standard step-indexed GRU without access to $\Delta t_t$
does not enforce this elapsed-time dependence. In contrast, TFP explicitly
parameterizes its retention as
\begin{equation}
k_t
=
\exp(-\Delta t_t/\tau_t),
\label{eq:ltc_retention_comparison}
\end{equation}
so the belief update remains calibrated to the physical interval between
policy queries.

This distinction does not imply that GRUs or LSTMs are incapable of
representing temporal history. Rather, elapsed-time consistency must be
learned from additional timing inputs or introduced through architectural
constraints, whereas it is built directly into the TFP update.

\subsection{Relation to Fixed-Decay Memory}
\label{sec:fixed_decay}

A fixed-decay memory updates its state according to
\begin{equation}
h_t
=
\exp(-\Delta t_t/\tau_0)\odot h_{t-1}
+
\left(1-\exp(-\Delta t_t/\tau_0)\right)\odot \hat{h}_t,
\label{eq:fixed_decay}
\end{equation}
where $\tau_0$ is a constant time scale. This model is a special case of the
TFP update obtained by setting $\tau_t=\tau_0$. TFP instead predicts
\begin{equation}
\tau_t
=
\tau_\theta(x_t,h_{t-1}),
\label{eq:adaptive_time_constant}
\end{equation}
allowing the effective retention time to depend on both the current evidence
and the retained belief. Consequently, TFP preserves the elapsed-time
structure of exponential retention while adapting how quickly individual
memory channels update across different task states and interaction events.

\subsection{Relation to State-Space Models}
\label{app:ssm_comparison}

Linear state-space models provide efficient long-horizon sequence modeling
through structured linear recurrences~\cite{gu2021s4,gu2023mamba}. Continuous-
time SSMs can also implement elapsed-time-dependent transitions, for example
\[
h_t=\exp(A\Delta t_t)h_{t-1}+B(\Delta t_t)x_t.
\]
Therefore, we do not claim that LTC is more time-consistent than all SSMs.
Instead, the distinction is functional: TFP uses LTC as a nonlinear,
input-dependent belief fuser whose state directly conditions a generative action
decoder. This makes the memory update both time-calibrated and action-relevant
for chunked manipulation policies.

\subsection{Computational Overhead}
\label{app:computational_overhead}

The LTC update costs $\mathcal{O}(d_h(d_x+d_h))$, where $d_x$ is the compact
observation dimension and $d_h$ is the memory dimension. The memory projection
costs $\mathcal{O}(d_hd_c)$, and AdaLN adds only lightweight affine modulation in
the action decoder. These costs are small relative to the vision-language
backbone and transformer-based action decoder; the main computational cost of
TFP comes from recurrent training over multiple contiguous chunks.

\begin{figure*}[t]
    \centering
    \includegraphics[width=\textwidth]{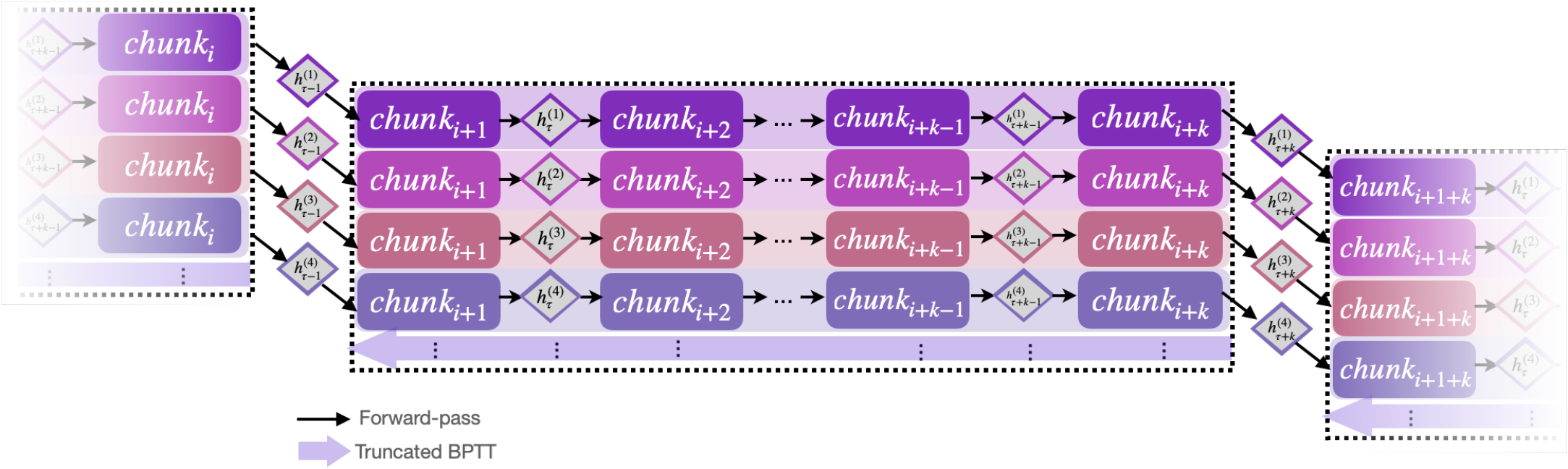}
    \caption{Episode-Aware Temporal Batching trains on contiguous chunks while carrying episode-local hidden states across truncated segments.}
    \label{fig:training}
\end{figure*}

%% file: sections/Experiments.tex
\section{Experiments}
\label{sec:experiments}
\raggedbottom

\noindent\textbf{Evaluation Questions.}
Our experiments ask whether TFP (i) improves standard and robustness-oriented manipulation benchmarks, (ii) improves real-world stage-dependent manipulation where the current observation alone is insufficient, (iii) encodes action-relevant task progress that causally affects action generation, and (iv) benefits from continuous-time LTC dynamics and real elapsed-time conditioning under irregular control intervals.

\paragraph{Experimental Setup.}

We evaluate TFP with $\pi_{0.5}$ as the primary VLA backbone on LIBERO, LIBERO-plus, MIKASA-Robo, and two Galaxea A1 real-world tasks: object swapping and counting pick-and-place. For our controlled same-backbone experiments, all reproduced variants use the same demonstrations, action horizon, optimizer, and rollout protocol. Published baselines are reported as literature references when available. Across experiments, TFP receives a language prompt, two RGB observations, and proprioceptive state at each policy query, and maintains an episode-local 256-dimensional memory state \(h_t\).
For recurrent methods, EATB is applied during training, memory states are episode-local and
reset only at episode boundaries during inference.

For comparisons to published baselines, we report their original fixed-horizon
numbers when available and mark them as literature references. For our controlled
same-backbone ablations, including recurrent variants, TFP
w/o \(\Delta t\), and TFP, we use the same adaptive receding-horizon executor.
Thus, policy queries occur after variable executed prefixes, and all controlled
memory variants are evaluated under the same irregular-query schedule. The only
difference between TFP and TFP w/o \(\Delta t\) is whether the memory update
receives the measured elapsed interval or a constant step interval. The real-world setup uses a
single-arm Galaxea A1 platform with a parallel gripper, a wrist-mounted fisheye
camera, and a static Intel RealSense D435 camera. Predicted action chunks are
executed as joint-angle commands through ROS1 Noetic.

Statistical reliability for the reproduced LIBERO and real-world results is reported in Appendix~\ref{app:statistical-reliability}.

\subsection{Main Benchmark Results}
\label{sec:main_benchmark_results}

We organize our evaluation around different benchmark purposes. LIBERO and LIBERO-plus measure online rollout performance and general manipulation quality, including long-horizon and multi-stage tasks. This is important because real-world manipulation is not always explicitly memory-dependent; a memory-augmented VLA should improve memory-sensitive behavior without sacrificing general task execution. Tables~\ref{tab:libero_main_results} summarizes the results on simulation benchmarks. 

We then use memory-focused benchmarks and real-world experiments as sanity checks for whether the proposed memory state is used. MIKASA-Robo ShellGameTouch isolates belief-state filtering under occlusion in simulation, while our real-world A1 experiments evaluate memory-dependent manipulation on hardware. Thus, LIBERO benchmark tests generality and rollout quality, whereas MIKASA-Robo and real-world A1 experiments test the intended memory mechanism, and LIBERO-plus further evaluates zero-shot generality under occlusion and visual perturbations, where the belief state is expected to stabilize action generation when the current observation is incomplete or unreliable. Ablations with recurrent, SSM-based variants are reported in Sec.~\ref{sec:ablations}.

\begin{table*}[t]
\centering
\caption{Success rate (\%) on LIBERO, LIBERO-plus, and MIKASA-Robo ShellGameTouch. TFP obtains the best average success rate on both LIBERO and LIBERO-plus while remaining in the 3.3B-parameter-only regime. Numbers for prior methods are reported results from
published work and are not necessarily reproduced under our training pipeline. We use ShellGameTouch as a diagnostic rather than a state-of-the-art comparison.}
\label{tab:libero_main_results}
\label{tab:libero_plus_main_results}
{\scriptsize\setlength{\tabcolsep}{2.3pt}
\setlength{\dashlinedash}{1.2pt}
\setlength{\dashlinegap}{1.6pt}
\resizebox{\linewidth}{!}{%
\begin{tabular}{lccccc:cccc:c}
\toprule
& \multicolumn{5}{c:}{\textbf{LIBERO}} & \multicolumn{4}{c:}{\textbf{LIBERO-plus}} & \textbf{MIKASA-Robo} \\
\textbf{Policy} & \textbf{Spatial} & \textbf{Object} & \textbf{Goal} & \textbf{Long-10} & \textbf{Avg.} & \textbf{Noise} & \textbf{Light} & \textbf{Backgr.} & \textbf{Avg.} & \textbf{ShellGameTouch} \\
\midrule
$\pi_{0.5}$ (3.3B) & 98.8 & 98.0 & 98.2 & 92.4 & 96.85 & 85.2 & 93.9 & 95.1 & 91.4 & -- \\
$\pi_0$ (3.3B) & 94.2 & 96.4 & 96.8 & 87.6 & 93.8 & 79.0 & 85.0 & 81.4 & 81.8 & 33 \\
OpenVLA (7B) & 84.7 & 88.4 & 79.2 & 53.7 & 76.5 & 15.2 & 8.1 & 34.8 & 19.37 & 12 \\
OpenVLA-OFT (7.3B) & 97.6 & 98.4 & 97.9 & 94.5 & 97.1 & 75.8 & 88.7 & 93.3 & 85.93 & 47 \\
HAMLET (3B) & 99.0 & \best{100.0} & 99.2 & 92.2 & 97.6 & -- & -- & -- & -- & -- \\
AVA-VLA~\cite{xiao2025avavla} (7.3B) & 97.4 & 99.4 & 97.4 & \best{97.6} & 98.0 & 78.0 & 95.5 & 88.9 & 87.47 & -- \\
MemoryVLA (7.3B) & 98.4 & 98.4 & 96.4 & 93.4 & 96.65 & -- & -- & -- & -- & \best{88} \\
TFP (ours) (3.3B) & \best{99.6} & 99.0 & \best{99.4} & 97.0 & \best{98.75} & \best{88.5} & \best{96.1} & \best{96.7} & \best{93.77} & 75 \\
\bottomrule
\end{tabular}%
}
}
\end{table*}

\noindent\textbf{LIBERO.}
On LIBERO, TFP improves the average success rate from $96.9\%$ for the reactive $\pi_{0.5}$ baseline to $98.75\%$. The improvement is most visible on the \textit{long} split, where TFP reaches $97.0\%$ compared with $92.4\%$ for $\pi_{0.5}$. This supports our hypothesis that maintaining latent task progress is especially beneficial for long-horizon manipulation, where the current observation alone may be insufficient for selecting the correct action.

\noindent\textbf{LIBERO-plus.}
On LIBERO-plus, TFP achieves the best average robustness score of $93.77\%$, outperforming $\pi_{0.5}$ by $2.37$ percentage points. The gains are most notable under noise and lighting perturbations, where TFP improves from $85.2\%$ to $88.5\%$ and from $93.9\%$ to $96.1\%$, respectively. 

\noindent\textbf{MIKASA-Robo.}
On MIKASA-Robo ShellGameTouch, TFP achieves a success rate of 75.0\%,
substantially improving over the reported OpenVLA-OFT reference result at 47.0\%.
ShellGameTouch isolates belief-state filtering under occlusion: after the ball
is hidden by visually identical mugs, the policy must remember the hidden ball
location and use that belief to touch the correct mug. Since the Touch variant
minimizes additional pushing or grasping requirements, the improvement more
directly reflects memory-conditioned action selection rather than low-level
manipulation skill. We do not present this result as a state-of-the-art claim:
MemoryVLA reports 88.0\% on this task. Instead, we use ShellGameTouch as a
diagnostic showing that TFP improves hidden-state reasoning in a reactive VLA,
while the remaining gap points to a complementary need for object-centric
hidden-location binding; see Appendix~\ref{app:mikasa_failure}.

\noindent\textbf{Real-world A1.}
Table~\ref{tab:real_world_failure} reports real-robot success counts and dominant
failure types among unsuccessful trials.

\begin{table}[H]
\centering
\caption{Real-world Galaxea A1 results and failure analysis.}
\label{tab:real_world_failure}
\label{tab:real_world_Failure}
{\scriptsize
\setlength{\tabcolsep}{3.4pt}
\renewcommand{\arraystretch}{0.86}
\begin{tabular}{llccccc}
\toprule
\textbf{Task} & \textbf{Method} & \textbf{Success} &
\textbf{Stage} & \textbf{Repeat} & \textbf{Target} &
\textbf{Exec.} \\
\midrule
\multirow{2}{*}{Object swap}
& $\pi_{0.5}$ & $3/20$  & \(\checkmark\) & \(\checkmark\) & \(\circ\) & -- \\
& TFP         & $15/20$ & -- & -- & \(\circ\) & \(\checkmark\) \\
\midrule
\multirow{2}{*}{Counting pick-place}
& $\pi_{0.5}$ & $8/20$  & \(\checkmark\) & \(\checkmark\) & -- & \(\circ\) \\
& TFP         & $18/20$ & -- & \(\circ\) & \(\circ\) & \(\checkmark\) \\
\bottomrule
\end{tabular}

\vspace{0.15em}
\textit{Failure tags:} Stage = wrong phase/forgotten progress;
Repeat = completed subtask repeated; Target = wrong object/placement grounding;
Exec. = low-level grasp, placement, or contact instability.
}

\end{table}
Table~\ref{tab:real_world_failure} shows that TFP primarily reduces
stage-level memory failures: the reactive baseline often repeats completed
subtasks or switches to the wrong stage, whereas TFP failures shift toward
target grounding and low-level execution errors. This suggests that the learned
belief improves task-progress tracking rather than merely improving low-level
manipulation.

\FloatBarrier

\subsection{Event-Sensitive Belief Dynamics}
\label{sec:memory_dynamics}

\begin{figure*}[!t]
    \centering
    \includegraphics[width=0.95\textwidth]{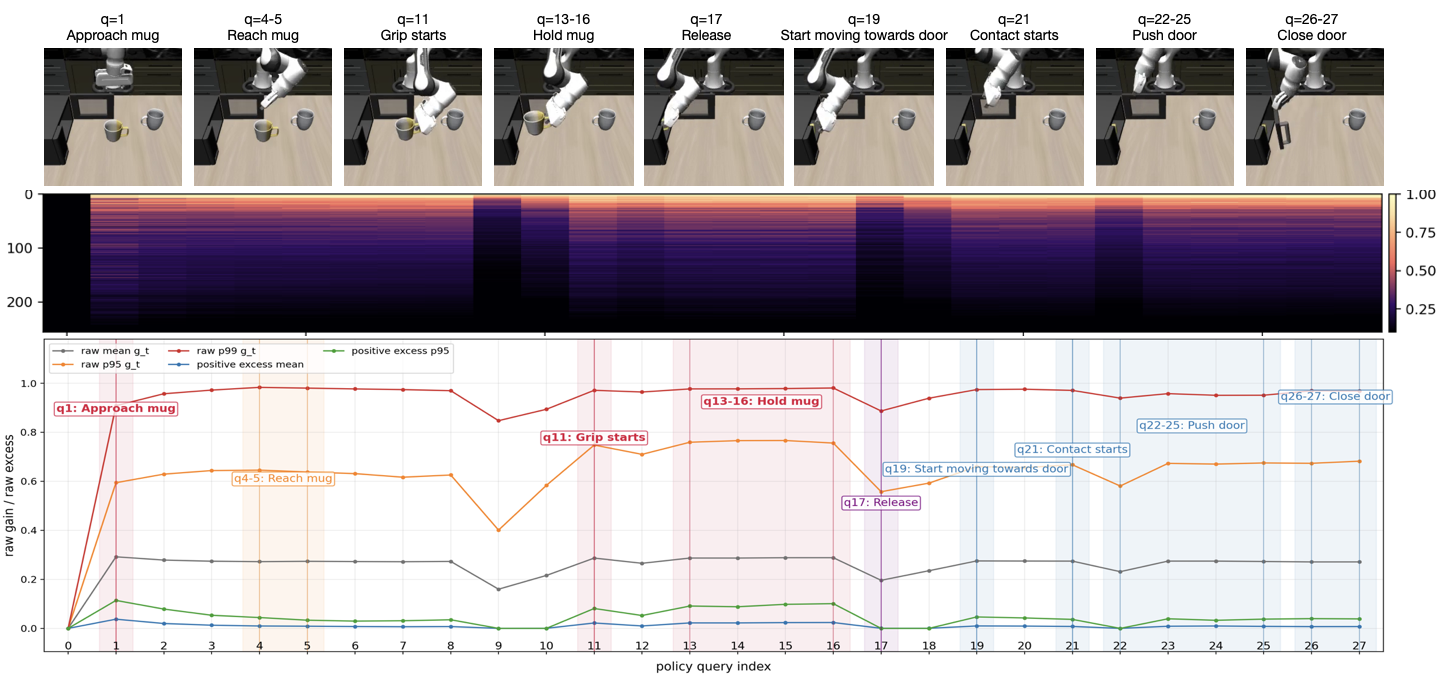}
    \vspace{0.2em}
    \caption{LTC belief update gain \(g_t\) during a rollout. Frames, dimension-wise gate heatmaps, and summary statistics show gate changes around reaching, carrying, releasing, and pushing events.}
    \label{fig:ltc_gain_dynamics}
\end{figure*}

\paragraph{Irregular-query diagnostic.}
We next examine whether the LTC memory learns an event-sensitive belief update
rather than a generic step-indexed recurrence. For this diagnostic, we run TFP
with adaptive receding-horizon execution: the policy predicts an action
chunk, but the executor may re-query after a variable executed prefix when the
chunk crosses a gripper transition, contact-sensitive phase, or unstable region.
The memory update therefore receives the elapsed interval between policy
queries. This setting exposes the memory module to irregular physical time, so
the learned update gain can reflect both manipulation events and elapsed time
rather than a fixed recurrent step index.

At each policy query, we record the internal LTC variables during LIBERO
rollouts: the time constant $\tau_t$, retention gate $k_t$, update gain
$g_t = 1-k_t$, hidden state $h_t$, and candidate belief $\hat h_t$. Larger
values of $g_t$ indicate stronger writing from the current observation, whereas
smaller values indicate stronger retention of the previous task belief. Thus,
the temporal pattern of $g_t$ directly reveals when the model chooses to revise
or preserve its latent belief.

For interpretation only, we extract privileged simulator states such as gripper state, end-effector position, object poses, object height, and
task success; these signals are never provided to the policy.




Figure~\ref{fig:ltc_gain_dynamics} shows nonuniform LTC updates: some dimensions
remain stable during transport, while others change near task transitions. To
quantify this event sensitivity, we measure write-gain changes
\(D^g_t=\frac{1}{d_h}\sum_i |g_{t,i}-g_{t-1,i}|\). For each simulator-side event,
we take the largest \(D^g_t\) within \(\pm 6\) queries and compare it with a far
non-event baseline excluding \(\pm 8\)-query neighborhoods. Table~\ref{tab:lagged_event_alignment}
shows that gain-pattern changes near events are about \(6\times\) larger than
the far non-event baseline, indicating that the policy often adjusts its
write/retain schedule slightly before the simulator-side event label.

\begin{table}[H]
\centering
\caption{Lagged alignment of LTC write-gain dynamics on LIBERO mug-in-microwave, using 224 events from 30 successful rollouts.}
\label{tab:lagged_event_alignment}
\scriptsize
\setlength{\tabcolsep}{2pt}
\resizebox{\columnwidth}{!}{%
\begin{tabular}{lcccc}
\toprule
\textbf{Metric} &
\textbf{Center/base} &
\textbf{Best $\pm6$/base} &
\textbf{95\% CI} &
\textbf{Lag} \\
\midrule
Mean dim. \(|\Delta g_t|\) &
\(0.42\times\) &
\(\mathbf{5.97\times}\) &
\([4.61, 7.54]\) &
\(-1.64\) \\
Top-10\% dim. \(|\Delta g_t|\) &
\(0.48\times\) &
\(5.90\times\) &
\([4.57, 7.44]\) &
\(-2.13\) \\
\(|\Delta \bar g_t|\) &
\(0.34\times\) &
\(6.24\times\) &
\([4.81, 7.88]\) &
\(-1.78\) \\
\(\|\Delta g_t\|_2/\sqrt{d_h}\) &
\(0.35\times\) &
\(6.14\times\) &
\([4.65, 7.88]\) &
\(-1.88\) \\
\bottomrule
\end{tabular}%
}
\end{table}

\noindent Following prior analyses of policy dynamics and task-relevant factors~\cite{wang2023measuring}, we use event labels only post hoc: they are not provided to the policy, but reveal whether LTC update gains track manipulation changes.

\paragraph{Same Observation, Different Memory States.}
Fixing the observation, robot state, and instruction while replacing only the LTC hidden state substantially changes the predicted 10-step action chunks: mean pairwise action distance is \(0.0442\), maximum \(0.1303\), while temporally adjacent hidden states have median distance \(0.0108\). Thus, the LTC belief actively conditions actions; see Appendix~\ref{app:hidden_state_intervention}.

\FloatBarrier
\subsection{Ablation Studies}
\label{sec:ablations}

\begin{table*}[!t]
\centering
\caption{
Controlled memory-dynamics ablations under the same adaptive receding-horizon
executor. All same-backbone variants use the same demonstrations, action horizon,
optimizer, rollout protocol, and irregular-query execution schedule. TFP w/o
\(\Delta t\) uses the same LTC and AdaLN architecture as TFP but replaces the
measured elapsed interval with a constant step size.
}
\label{tab:memory_dynamics_ablation}
\footnotesize
\renewcommand{\arraystretch}{1.08}
\setlength{\tabcolsep}{4.2pt}

\begin{minipage}{0.57\textwidth}
\centering
\textbf{(a) LIBERO~\cite{liu2023libero}}
\vspace{0.4em}

\begin{tabular}{lccccc}
\toprule
\textbf{Method} & \textbf{Spatial} & \textbf{Object} & \textbf{Goal} & \textbf{Long} & \textbf{Avg.} \\
\midrule
$\pi_{0.5}$              & 98.8 & 98.0 & 98.2 & 92.4 & 96.9 \\
$\pi_{0.5}$ + GRU        & 99.0 & 98.5 & 95.5 & 89.0 & 95.5 \\
$\pi_{0.5}$ + SSM (S4D)  & 98.5 & 98.5 & 97.0 & 92.5 & 96.625 \\
TFP w/o $\Delta t$       & 98.5 & 97.5 & 97.5 & 94.0 & 96.9 \\
TFP (ours)               & \best{99.6} & \best{99.0} & \best{99.4} & \best{97.0} & \best{98.75} \\
\bottomrule
\end{tabular}
\end{minipage}
\hfill
\begin{minipage}{0.40\textwidth}
\centering
\textbf{(b) LIBERO-plus~\cite{fei25libero-plus}}
\vspace{0.4em}

\begin{tabular}{lcccc}
\toprule
\textbf{Method} & \textbf{Noise} & \textbf{Light} & \textbf{Bg.} & \textbf{Avg.} \\
\midrule
$\pi_{0.5}$              & 85.2 & 93.9 & 95.1 & 91.4 \\
$\pi_{0.5}$ + GRU        & 85.9 & 91.5 & 96.1 & 91.2 \\
$\pi_{0.5}$ + SSM (S4D)  & 82.6 & 95.0 & 92.3 & 90.0 \\
TFP w/o $\Delta t$       & 82.5 & 93.8 & 94.1 & 90.1 \\
TFP (ours)               & \best{88.5} & \best{96.1} & \best{96.7} & \best{93.8} \\
\bottomrule
\end{tabular}
\end{minipage}

\end{table*}

We ablate whether TFP's gains come merely from adding temporal state, and whether elapsed-time conditioning in the LTC update is necessary.

\subsubsection{Memory Dynamics}
\label{sec:ablation_memory_dynamics}

We compare TFP with GRU and S4D memory variants under the same $\pi_{0.5}$ backbone, training data, action horizon, and LIBERO evaluation protocol. This tests whether gains come from generic recurrence or from continuous-time memory coupled to action-head conditioning.

GRU performs competitively on spatial tasks but degrades on object, goal, and long-horizon tasks; S4D improves over GRU but remains below TFP, especially on the long-horizon split. Thus, gains are not from memory alone: the state must be temporally stable and directly coupled to action generation.

\paragraph{Elapsed-time conditioning.}
Because adaptive receding-horizon execution re-queries the policy after variable
executed prefixes, the interval between memory updates is nonuniform. We isolate
the role of measured elapsed time by comparing TFP with TFP w/o \(\Delta t\),
which keeps the same LTC update, AdaLN action conditioning, training data, and
adaptive executor, but replaces the measured interval with a constant step size.
The drop from TFP to TFP w/o \(\Delta t\), especially on LIBERO Long and
LIBERO-plus, shows that the gain is not only from adding a recurrent state or
adaptive execution; the memory update benefits from receiving the actual elapsed
time between policy queries.

On LIBERO-plus, GRU improves some perturbations but lowers average robustness relative to $\pi_{0.5}$, while TFP improves the average from $91.4\%$ to $93.8\%$, mainly under noise and lighting. Memory must therefore stabilize action generation rather than merely add recurrent state.
\FloatBarrier
\subsection{Limitations}
\label{sec:limitations}

Although LTC and AdaLN add little inference overhead, recurrent full fine-tuning is costly because training must unroll across action chunks to preserve hidden-state continuity and credit assignment. EATB truncates gradients across contiguous chunks, but our largest LIBERO runs still used $K=8$, batch size 128, over 100GB GPU memory, and about 80 hours on 4 H200 GPUs to reach imitation loss near 0.003. Our real-world evaluation is also limited to tabletop single-arm manipulation, leaving mobile manipulators, humanoids, and dexterous hands for future work.

%% file: sections/Conclusion.tex
\FloatBarrier
\section{Conclusion}
\label{sec:conclusion}

We presented Temporally Conditioned Memory-Fusion Policies (TFP), a framework
for extending reactive VLA policies with an explicit task-progress belief for
history-dependent manipulation. TFP maintains this belief with Liquid
Time-Constant dynamics and injects it directly into the flow-matching action
decoder through AdaLN-style conditioning, allowing the learned belief to
shape action generation rather than serve as passive history context. Across
simulation and real-world tasks, TFP improves performance most clearly when the
current observation alone is insufficient, including long-horizon, visually
perturbed, occluded, and stage-dependent manipulation settings. Mechanistic
analyses show that the learned write-gain dynamics change sharply near
manipulation-event boundaries, while hidden-memory interventions show that
changing only the belief can alter the generated action chunk and end-effector
trajectory. These results suggest that memory in VLA policies should act as an
event-sensitive action-conditioning state: retaining task progress during stable
or ambiguous phases, and revising the belief when task-relevant evidence
changes. Future work should combine such dynamics-aware memory with stronger
object-centric grounding, spatial generalization, and compositional planning, while developing more efficient recurrent fine-tuning schemes.

%% file: sections/References.tex
\bibliographystyle{plainnat}
\bibliography{reference}

%% file: appendices/Appendix.tex
\appendices

\providecommand{\vecsym}[1]{\mathbf{#1}}

The appendix provides additional material omitted from the main paper for space.
Appendix~\ref{app:related_work} gives a broader discussion of related work.
Appendix~\ref{app:method_details} describes implementation and execution details.
Appendix~\ref{app:belief_interpretation} formalizes the continuous-time belief
interpretation of LTC memory. Appendix~\ref{app:baseline_comparison} compares
TFP with recurrent, fixed-decay, and state-space alternatives.
Appendix~\ref{app:experimental_details} provides additional experimental
details, including hidden-state interventions and MIKASA-Robo failure analysis.
Appendix~\ref{app:statistical-reliability} reports statistical reliability
results for simulation and real-world rollouts.

\section{Additional Related Work}
\label{app:related_work}

\paragraph{Vision-language-conditioned robot policies.}Robot manipulation has increasingly moved from task-specific behavior cloning toward vision-language-conditioned policies that map images, language instructions, and robot states to actions. RT-2, OpenVLA, Octo, and $\pi_0$ extend this direction to generalist VLA policies trained with large-scale robot and vision-language data~\cite{brohan2023rt2,openvla2024,octo2024,black2024pi0}. Related architectures such as PerAct further show the benefit of structured visual representations for language-conditioned manipulation~\cite{shridhar2023perceiver}. TFP builds on this family of policies, but targets a different failure mode: in stage-dependent manipulation, the current observation and instruction may not uniquely determine the correct action without an internal belief over task progress, hidden object state, or previous interaction outcomes.

\paragraph{Chunked generative action policies.}Recent imitation policies often predict action chunks rather than single-step actions. ACT learns sequence-level action prediction for fine-grained manipulation, Diffusion Policy represents visuomotor control as conditional denoising over action trajectories, and $\pi_0$ uses flow matching for generalist action generation~\cite{zhao2023learning,chi2023diffusion,black2024pi0}.Action chunking improves temporal consistency and reduces policy-query frequency, but it also creates a memory problem: after executing part of a chunk, the policy must decide what task-progress information should persist to the next query. TFP addresses this by maintaining a latent belief across policy queries and injecting it directly into the action decoder.

\paragraph{Temporal and continuous-time memory models.}Temporal context can be modeled with recurrent networks, self-attention,state-space models, and continuous-time recurrent dynamics. RNNs, LSTMs, and GRUs provide compact recurrent states for history aggregation, with LSTMs designed for long time-lag dependencies and GRUs offering a gated recurrence~\cite{6795963,cho2014learning}. Time-aware variants further incorporate elapsed time through decay mechanisms or temporal gates, including GRU-D,T-LSTM, DATA-GRU, and Phased LSTM~\cite{che2018recurrent,baytas2017patient,tan2020data,neil2016phased}.These models are natural baselines for history-dependent control, but in chunked visuomotor policies they still require sequential unrolling over policy queries. Because each recurrent step may involve a large VLA backbone, an action-generation decoder, and a full action-horizon imitation loss, repeatedly calling the policy across many memory steps can substantially increase training cost and limit temporal parallelism.

Self-attention provides parallel content-based access to history, but using past observations and action chunks as an ever-growing token set increases context length and does not by itself impose a physical-time belief dynamics~\cite{vaswani2017attention}. Structured state-space models such as S4 and selective SSMs such as Mamba offer efficient long-sequence modeling with linearor near-linear scaling~\cite{gu2021s4,gu2023mamba}. These methods are strong sequence encoders, but our setting requires more than long-context compression:the policy must maintain an online task-progress belief whose retention is calibrated by elapsed time and whose state directly conditions action generation.

Continuous-time recurrent models provide another route to temporal memory.Classical continuous-time recurrent neural networks model hidden-state evolution with differential equations, while Neural ODEs, ODE-RNNs, and LatentODEs evolve hidden states continuously between irregular observations~\cite{funahashi1993approximation,chen2018neuralode,rubanova2019latent}.Liquid Time-Constant networks and closed-form continuous-time networks further parameterize input-dependent time constants for adaptive continuous-time sequence modeling~\cite{hasani2021ltc,hasani2022cfc}. TFP builds on this continuous-time memory family, but uses LTC not as a standalone sequence model:we use it as a compact belief filter for chunked VLA control, with vector-valued, input-dependent time constants and closed-form elapsed-time retention. The resulting memory is then projected into the action-head conditioning space, so temporal belief directly modulates the generated action chunk instead of remaining a passive history representation.

\paragraph{Memory-augmented and history-aware visuomotor policies.}Several recent works augment robot policies with history or memory to address partial observability and long-horizon manipulation. HAMLET adapts pretrained VLAs into history-aware policies by introducing moment tokens and a lightweight memory module that aggregates historical context for action prediction~\cite{koo2025hamlet}. This provides an efficient alternative to naively appending multiple past frames, but its memory primarily serves as a token-level history compression and retrieval interface. MemoryVLA introduces a cognition-memory-action framework, where perceptual and cognitive tokens form working memory, retrieve decision-relevant entries from a memory bank, and condition a diffusion action expert~\cite{shi2025memoryvla}. This design preserves richer historical details, but the temporal state is organized around memory storage, retrieval, and consolidation rather than an explicit elapsed-time-conditioned task dynamics. AVA-VLA reformulates VLA control from a POMDP perspective and uses a recurrent belief state to actively modulate visual token processing~\cite{xiao2025avavla}. However, its memory is mainly used to guide perception, whereas the action-generation pathway is not explicitly conditioned by a continuous-time task-progress belief.

History-aware diffusion policies are also related. Causal Diffusion Policy conditions diffusion-based visuomotor control on historical action sequences through a causal transformer, improving temporal coherence and robustness under degraded observations~\cite{cdp2025}. Its focus is autoregressive action history modeling rather than language-conditioned VLA control, and its own limitation notes that extremely long-horizon tasks requiring sustained planning remain unresolved. These methods show that exposing the policy to better,richer, or more retrievable history is valuable. TFP addresses a complementary problem: the policy should not only access history, but should maintain a compact latent belief whose evolution follows temporal-conditioned task dynamics.

Concretely, TFP compresses history into a fixed-size continuous-time belief state with input-dependent LTC retention and injects this belief directly into the generative action decoder through AdaLN-style modulation. Thus, memory is not only a retrieved context or an auxiliary visual-attention signal; it directly modulates the decoder activations and therefore the flow-matching velocity field over the predicted action chunk. This makes TFP better aligned with real manipulation dynamics, where the correct action depends on how task progress evolves over physical time, not merely on whether the model can attend to a more complete history.

\section{Additional Method Details}
\label{app:method_details}

\subsection{Memory-Conditioning Interface}
\label{app:memory_conditioning_interface}

TFP conditions the action decoder through adaptive modulation rather than
explicit memory-token cross-attention. Cross-attention-based memory injection
requires the decoder to retrieve and interpret a separate set of memory tokens.
In contrast, adaptive conditioning compresses temporal context into a single
modulation vector that directly shapes decoder activations. The timestep
embedding captures where the denoising process is, while the memory condition
captures what temporal context is relevant. Their additive fusion lets the
decoder jointly adapt to diffusion phase and retained task belief without adding
an extra attention pathway.

This choice reflects the role of LTC memory as a compact time-consistent belief
state rather than a retrievable history buffer. We do not claim that
cross-attention is inherently inferior; instead, AdaLN provides a simpler
fixed-size interface for injecting belief into every action-decoder layer. A
computational overhead discussion is provided in
Appendix~\ref{app:computational_overhead}.

\subsection{Episode-Aware Temporal Batching Details}
\label{app:eatb_details}

For each episode $e$, EATB keeps an episode-local hidden state
$\vecsym{h}_t^{(e)}=f_\theta(\vecsym{h}_{t-1}^{(e)},o_t^{(e)},\Delta t_t^{(e)})$,
with $\vecsym{h}_0^{(e)}=\mathbf{0}$, and resets it only at episode boundaries.
At training step $s$, we sample $B$ active episodes and unroll each for $K$
consecutive chunks. If episode $e_b$ starts from chunk index $\tau_b$, its
stored hidden state is retrieved as
$\vecsym{h}_{b,0}=\tilde{\vecsym{h}}_{\tau_b}^{(e_b)}$ and updated as
\begin{equation}
\vecsym{h}_{b,k}
=
f_\theta(\vecsym{h}_{b,k-1},\vecsym{x}_{b,k},\Delta t_{b,k}),
\qquad k=1,\dots,K .
\end{equation}
The optimization objective is
\begin{equation}
\mathcal{L}_s
=
\frac{1}{BK}
\sum_{b=1}^{B}
\sum_{k=1}^{K}
\ell_{\mathrm{flow}}^{(b,k)} .
\end{equation}
After the parameter update, the final hidden state is written back with gradient
truncation,
\begin{equation}
\tilde{\vecsym{h}}_{\tau_b+K}^{(e_b)}
\leftarrow
\operatorname{stopgrad}(\vecsym{h}_{b,K}) .
\end{equation}
This preserves forward memory across training segments while bounding the
backpropagation length.

\section{Continuous-Time Belief Interpretation}
\label{app:belief_interpretation}

This section formalizes the connection between Liquid Time-Constant (LTC) memory
and belief filtering in partially observable manipulation. We do not claim that
LTCs are universally more expressive than all recurrent models. A sufficiently
large recurrent network with access to elapsed time may approximate many
belief-update functions. Instead, our claim is more specific: LTC provides a
structural inductive bias that matches a class of continuous-time,
multi-timescale belief dynamics. This class is particularly relevant for
chunked robot manipulation, where policy queries may occur after variable
execution intervals.

\paragraph{Claim.}
For manipulation tasks that can be modeled as partially observable decision
processes with latent task state, the desired memory should behave as a
time-consistent belief state. LTC implements the exact discretization of a
first-order continuous-time belief relaxation process, and its exponential
retention gate is the unique continuous scalar retention rule satisfying
elapsed-time consistency. Step-indexed recurrent baselines such as GRUs or
LSTMs do not enforce this property unless elapsed time is explicitly provided
and the gate is constrained to satisfy the same exponential semigroup structure.

\subsection{Belief Filtering in a Continuous-Time POMDP}
\label{app:pomdp_belief}

Let \(z_t\) denote the latent task state, which may include task phase, object
state, object-in-hand status, subgoal completion, and other variables not fully
observable from the current image. Let \(o_t\) denote the observation, including
visual and proprioceptive inputs. In a partially observable Markov decision
process, the sufficient statistic for optimal decision making is the belief
state
\begin{equation}
b_t(z)
=
p(z_t=z \mid o_{1:t}, a_{1:t-1}).
\end{equation}
The standard discrete-time Bayes filter updates the belief by combining a
prediction step with an observation likelihood:
\begin{equation}
b_t(z_t)
=
\eta\,
p(o_t \mid z_t)
\sum_{z_{t-1}}
p(z_t \mid z_{t-1}, a_{t-1})
b_{t-1}(z_{t-1}),
\label{eq:discrete_bayes_filter}
\end{equation}
where \(\eta\) is a normalization constant.

In continuous time, if the latent Markov dynamics have generator \(Q\), the
prediction step over an elapsed interval \(\Delta t_t\) is
\begin{equation}
\bar b_t
=
\exp(Q \Delta t_t)b_{t-1}.
\label{eq:ct_belief_prediction}
\end{equation}
Thus, belief evolution depends on elapsed physical time, not only on the number
of policy queries. This motivates memory updates whose retention is explicitly a
function of \(\Delta t_t\).

In TFP, we do not represent the full probabilistic belief \(b_t\). Instead, the
memory state \(h_t \in \mathbb{R}^{d_h}\) is a learned belief embedding that
summarizes task progress and interaction history.

\subsection{LTC as First-Order Belief Relaxation}
\label{app:ltc_discretization}

Assume that, at query \(t\), the current observation and prior memory induce a
candidate belief
\begin{equation}
\hat h_t
=
C_\theta(x_t,h_{t-1}),
\end{equation}
where \(x_t\) is the compact observation representation. Consider the following
continuous-time relaxation dynamics for each latent belief coordinate \(j\):
\begin{equation}
\frac{d h_j(u)}{d u}
=
-\frac{1}{\tau_{t,j}}
\left(h_j(u)-\hat h_{t,j}\right),
\qquad
u \in [t-\Delta t_t,t],
\label{eq:belief_ode}
\end{equation}
where \(\tau_{t,j}>0\) is the time constant for memory channel \(j\). We assume
that \(\hat h_t\) and \(\tau_t\) are held fixed over the interval between two
policy queries.

\paragraph{Proposition 1.}
The exact solution of Eq.~\eqref{eq:belief_ode} over elapsed time
\(\Delta t_t\) is
\begin{equation}
h_{t,j}
=
\exp(-\Delta t_t/\tau_{t,j})h_{t-1,j}
+
\left(1-\exp(-\Delta t_t/\tau_{t,j})\right)\hat h_{t,j}.
\label{eq:scalar_ltc_solution}
\end{equation}
Equivalently, in vector form,
\begin{equation}
h_t
=
k_t \odot h_{t-1}
+
(1-k_t)\odot \hat h_t,
\qquad
k_t=\exp(-\Delta t_t/\tau_t).
\label{eq:ltc_solution}
\end{equation}

\paragraph{Proof.}
For each dimension \(j\), Eq.~\eqref{eq:belief_ode} is a linear first-order ODE:
\[
\frac{d h_j(u)}{d u}
=
-\frac{1}{\tau_{t,j}}h_j(u)
+
\frac{1}{\tau_{t,j}}\hat h_{t,j}.
\]
Let the initial value at the beginning of the interval be
\(h_j(t-\Delta t_t)=h_{t-1,j}\). Solving the ODE gives
\[
h_j(u)-\hat h_{t,j}
=
\exp\!\left(-\frac{u-(t-\Delta t_t)}{\tau_{t,j}}\right)
\left(h_{t-1,j}-\hat h_{t,j}\right).
\]
Setting \(u=t\) yields
\[
h_{t,j}-\hat h_{t,j}
=
\exp(-\Delta t_t/\tau_{t,j})
\left(h_{t-1,j}-\hat h_{t,j}\right).
\]
Rearranging gives Eq.~\eqref{eq:scalar_ltc_solution}. Applying this
coordinate-wise gives Eq.~\eqref{eq:ltc_solution}. \(\square\)

This shows that the LTC update is not an arbitrary interpolation rule: it is the
exact discrete-time update of a continuous-time belief relaxation process. The
candidate \(\hat h_t\) acts as evidence induced by the current observation, the
previous memory \(h_{t-1}\) acts as the retained prior belief, and the
time-dependent retention \(k_t\) determines how strongly the prior persists over
the elapsed interval.

\subsection{Elapsed-Time Consistency of Exponential Retention}
\label{app:elapsed_time_consistency}

The exponential form in Eq.~\eqref{eq:ltc_solution} is also the natural form
required by elapsed-time consistency. Consider a scalar memory channel whose
retention over elapsed time \(\Delta\) is \(k(\Delta)\). If the same interval is
split into two sub-intervals, \(\Delta_1\) and \(\Delta_2\), then applying
retention over \(\Delta_1\) and then over \(\Delta_2\) should be equivalent to
applying it once over \(\Delta_1+\Delta_2\). Therefore, a time-consistent
retention function must satisfy
\begin{equation}
k(\Delta_1+\Delta_2)
=
k(\Delta_1)k(\Delta_2).
\label{eq:semigroup}
\end{equation}

\paragraph{Proposition 2.}
Assume \(k:[0,\infty)\rightarrow(0,1]\) is continuous, satisfies \(k(0)=1\),
and obeys Eq.~\eqref{eq:semigroup}. Then
\begin{equation}
k(\Delta)=\exp(-\lambda \Delta)
\end{equation}
for some \(\lambda \ge 0\). Equivalently,
\[
k(\Delta)=\exp(-\Delta/\tau)
\]
for \(\tau=1/\lambda\), with \(\tau=\infty\) when \(\lambda=0\).

\paragraph{Proof.}
Since \(k(\Delta)>0\), define \(r(\Delta)=\log k(\Delta)\). Taking logarithms of
Eq.~\eqref{eq:semigroup} gives
\[
r(\Delta_1+\Delta_2)=r(\Delta_1)+r(\Delta_2).
\]
Because \(k\) is continuous, \(r\) is continuous. The continuous solutions to
Cauchy's additive functional equation on \([0,\infty)\) are linear:
\[
r(\Delta)=c\Delta.
\]
Since \(k(\Delta)\in(0,1]\), we have \(r(\Delta)\le 0\), so \(c=-\lambda\) for
some \(\lambda\ge 0\). Thus,
\[
k(\Delta)=\exp(r(\Delta))=\exp(-\lambda\Delta).
\]
\(\square\)

Thus, exponential retention is not an arbitrary design choice. It is the unique
continuous scalar retention rule that is consistent with elapsed-time
composition. LTC parameterizes this retention with an input-dependent time
constant,
\[
\tau_t=\tau_\theta(x_t,h_{t-1}),
\]
allowing each memory channel to adapt its effective time scale to the current
belief and evidence.

\subsection{Belief Stabilization Is Not Memory Collapse}
\label{app:belief_stabilization}

The LTC update also explains why low update gain does not necessarily indicate
memory degeneration. Suppose that within a manipulation stage, the relevant
latent belief is approximately constant:
\[
z_t=z.
\]
Assume that the candidate belief produced from the current observation is noisy:
\begin{equation}
\hat h_t=z+\epsilon_t,
\qquad
\mathbb{E}[\epsilon_t]=0,
\qquad
\operatorname{Var}(\epsilon_t)=\sigma^2.
\end{equation}
For a scalar memory channel with constant retention \(k\), the update is
\begin{equation}
h_t=kh_{t-1}+(1-k)(z+\epsilon_t).
\label{eq:noisy_filter}
\end{equation}

\paragraph{Proposition 3.}
Under the assumptions above and with \(h_0=0\), the mean and variance of \(h_t\)
are
\begin{equation}
\mathbb{E}[h_t]=(1-k^t)z,
\end{equation}
and
\begin{equation}
\operatorname{Var}(h_t)
=
(1-k)^2\sigma^2
\frac{1-k^{2t}}{1-k^2}.
\end{equation}
For large \(t\),
\begin{equation}
\operatorname{Var}(h_t)
\approx
\sigma^2\frac{1-k}{1+k}.
\label{eq:variance_limit}
\end{equation}

\paragraph{Proof.}
Unrolling Eq.~\eqref{eq:noisy_filter} gives
\[
h_t
=
k^t h_0
+
(1-k)
\sum_{i=1}^{t}
k^{t-i}(z+\epsilon_i).
\]
With \(h_0=0\),
\[
h_t
=
(1-k)
\sum_{i=1}^{t}
k^{t-i}z
+
(1-k)
\sum_{i=1}^{t}
k^{t-i}\epsilon_i.
\]
Since \(\mathbb{E}[\epsilon_i]=0\),
\[
\mathbb{E}[h_t]
=
(1-k)
\sum_{i=1}^{t}
k^{t-i}z
=
(1-k)(1+k+\cdots+k^{t-1})z.
\]
Using the geometric series identity,
\[
1+k+\cdots+k^{t-1}
=
\frac{1-k^t}{1-k},
\]
we obtain
\[
\mathbb{E}[h_t]=(1-k^t)z.
\]
For the variance, only the noise term contributes:
\[
\operatorname{Var}(h_t)
=
\operatorname{Var}
\left[
(1-k)\sum_{i=1}^{t}k^{t-i}\epsilon_i
\right].
\]
Assuming the noise terms are independent with variance \(\sigma^2\),
\[
\operatorname{Var}(h_t)
=
(1-k)^2
\sum_{i=1}^{t}
k^{2(t-i)}\sigma^2.
\]
The remaining sum is
\[
\sum_{i=1}^{t}k^{2(t-i)}
=
1+k^2+\cdots+k^{2(t-1)}
=
\frac{1-k^{2t}}{1-k^2}.
\]
Thus,
\[
\operatorname{Var}(h_t)
=
(1-k)^2\sigma^2
\frac{1-k^{2t}}{1-k^2}.
\]
For large \(t\), \(k^{2t}\rightarrow0\) when \(|k|<1\), and since
\(1-k^2=(1-k)(1+k)\), we get
\[
\operatorname{Var}(h_t)
\approx
\sigma^2\frac{1-k}{1+k}.
\]
\(\square\)

Equation~\eqref{eq:variance_limit} shows that when the candidate belief is noisy
but the underlying task belief is stable, larger retention \(k\) reduces the
variance of the memory state. Therefore, a high-retention regime can represent
belief stabilization rather than collapse. In manipulation, this is desirable
when slow variables such as task progress, target identity, or object-in-hand
status should persist across occlusion, contact noise, and action chunks.

\section{Additional Experimental Details}
\label{app:experimental_details}
\subsection{Same-Observation Hidden-State Intervention}
\label{app:hidden_state_intervention}

To isolate the effect of the LTC memory from perception, we conduct a controlled hidden-state intervention. We select one query from a LIBERO rollout and fix its observation, robot state, and language instruction. We then evaluate the policy multiple times while replacing only the LTC hidden state \(h_t\) with hidden states collected from 25 different rollout timesteps. All other inputs are kept unchanged. For each hidden condition, the policy predicts a 10-step action chunk in the environment action space.

Given two predicted action chunks \(A_i, A_j \in \mathbb{R}^{H \times d_a}\), we compute their normalized L2 distance:
\[
D(i,j) = \frac{1}{H}\sum_{r=1}^{H} \| A_{i,r} - A_{j,r} \|_2 .
\]
We report pairwise distances across all hidden-state pairs, distances to a reference hidden state, and distances between temporally adjacent hidden states. Since the observation, robot state, and instruction are fixed, any action variation is caused by the memory intervention.

\begin{figure}[t]
\centering
\begin{minipage}[t]{0.47\linewidth}
\vspace{0pt}
\centering
\captionof{table}{Hidden-state intervention under a fixed observation.}
\label{tab:hidden_state_intervention}
\small
\setlength{\tabcolsep}{4pt}
\begin{tabular}{l c}
\toprule
Metric & Value \\
\midrule
Hidden conditions & 25 \\
Pairwise comparisons & 300 \\
Mean pairwise dist. & 0.0442 \\
Median pairwise dist. & 0.0363 \\
Max pairwise dist. & 0.1303 \\
Ref. distance, mean & 0.0324 \\
Ref. distance, max & 0.1209 \\
Adjacent dist., median & 0.0108 \\
\bottomrule
\end{tabular}
\end{minipage}
\hfill
\begin{minipage}[t]{0.49\linewidth}
\vspace{0pt}
\centering
\includegraphics[width=\linewidth]{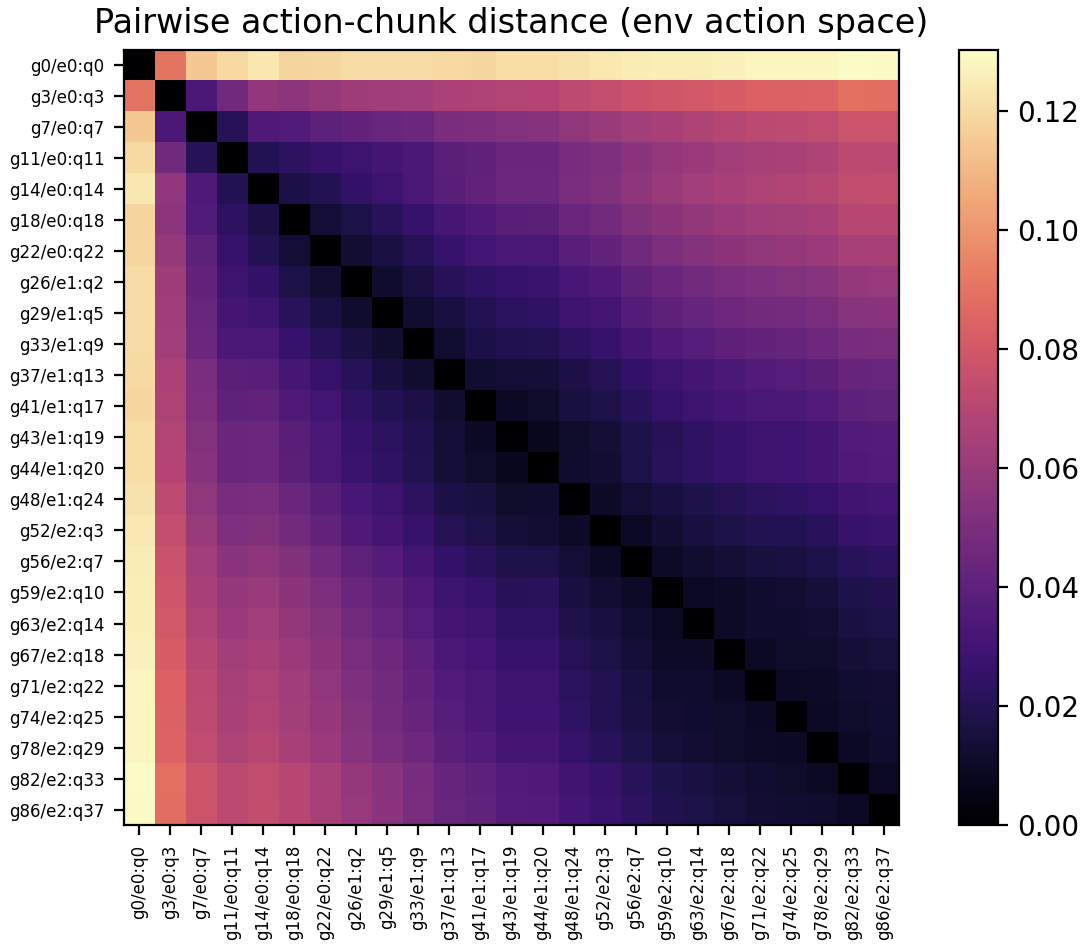}
\caption{Action-chunk variation caused by changing only the LTC hidden state under the same observation.}
\label{fig:hidden_state_intervention}
\end{minipage}
\end{figure}

\subsection{Failure Modes on MIKASA-Robo ShellGameTouch}
\label{app:mikasa_failure}

MIKASA-Robo ShellGameTouch provides a useful diagnostic for testing whether a
policy can preserve information that is no longer visible in the current
observation. In our same-backbone evaluation, TFP achieves the success rate of 75.0\%, indicating that the learned belief
state provides useful memory under occlusion. However, the remaining failures
also reveal an important boundary of the proposed method: ShellGameTouch
requires a memory regime that is only partially aligned with the event-sensitive
manipulation dynamics targeted by TFP.

\paragraph{A complementary memory regime.}
TFP is designed to model continuous task-progress dynamics in manipulation. Its
LTC belief update is most useful when the policy must decide when to preserve
slow task variables and when to revise them around interaction events such as
contact, grasp, release, placement, or subgoal completion. ShellGameTouch has a
different structure. The decisive variable is a discrete hidden object-location
binding: which visually identical mug hides the ball. Once the ball is occluded,
the current observation provides little corrective evidence about this latent
identity-location association. The policy must therefore write the correct
binding early and retrieve it at the final touch decision. This is closer to
object-centric episodic memory or key-value binding than to continuously
evolving manipulation-stage dynamics.

\paragraph{Sparse opportunities for belief correction.}
In many manipulation tasks, the belief can be refined as new contact, motion,
object-state, or subgoal-completion evidence appears. For example, after a
grasp, release, or placement event, the observation gives new evidence about
task progress, and an event-sensitive memory update can revise the internal
state accordingly. In ShellGameTouch, by contrast, the crucial information is
available only before and during occlusion. After the mugs become visually
ambiguous, later observations mainly support execution rather than correction
of the hidden ball-location belief. As a result, a weak or incorrect initial
memory write is difficult to repair through subsequent LTC updates.

\paragraph{Representation mismatch.}
This failure mode can also be viewed through a representation-learning lens.
Prior work on interpreting neural robot policies argues that useful policy
representations often align with task-relevant factors of variation, such as
skills, behaviors, or strategies~\cite{wang2023measuring}. Under this view,
ShellGameTouch stresses a different dominant factor of variation from the one
most directly targeted by TFP. The task-relevant factor in ShellGameTouch is a
persistent categorical object-location variable, whereas the factor targeted by
TFP is event-sensitive task-progress dynamics. TFP can still improve
ShellGameTouch because its belief state preserves useful latent context, but
the memory is not explicitly structured as a symbolic object map, slot memory,
or key-value store over hidden cup identities.

\paragraph{Failure categories.}
Qualitatively, we observe four main failure modes. First, the policy may fail
to encode the correct ball--mug association before occlusion, especially when
the relevant visual cue is brief or spatially small. Second, the memory may
retain a coarse task-progress signal but not a sufficiently precise
object-location binding, causing the final touch to target the wrong mug.
Third, the retained belief may be correct, but the action decoder may fail to
ground that belief into the correct end-effector target under visually similar
mugs. Finally, some failures are low-level execution errors near the final
target rather than clear memory errors. These categories suggest that
ShellGameTouch requires both persistent memory and precise object-centric
grounding.

\paragraph{Interpretation.}
We therefore treat ShellGameTouch as a same-backbone diagnostic rather than a
direct state-of-the-art comparison. The improvement over the reactive
baselines shows that TFP provides useful memory under occlusion, but
the absolute success rate reflects the mismatch between ShellGameTouch's
categorical hidden-location memory and TFP's primary design target:
continuous-time, event-sensitive manipulation belief dynamics. A natural future
direction is to combine TFP with an object-centric memory module, such as a
slot-based belief map or a retrieval mechanism for discrete hidden-state
bindings. Such a hybrid would allow the policy to retain symbolic
object-location facts while still using LTC dynamics to update task-progress
belief around manipulation events.

\section{Statistical Reliability of Rollout Results}
\label{app:statistical-reliability}

Because several LIBERO splits are close to saturation, we report confidence intervals for reproduced rollout results. For standard LIBERO, each suite contains 10 tasks and each task is evaluated with 50 rollout trials, giving 500 rollouts per suite. For each success rate, we compute a Wilson 95\% confidence interval from the number of successful rollouts. For method differences on standard LIBERO, we report approximate two-proportion 95\% confidence intervals against the reproduced $\pi_{0.5}$ baseline. We report intervals only for our reproduced methods; published prior-method numbers in Table~\ref{tab:libero_main_results} are not necessarily evaluated under the same rollout protocol.

\begin{table}[!htbp]
\centering
\small
\caption{Statistical reliability of reproduced LIBERO rollout results. Each LIBERO suite contains 10 tasks, and each task is evaluated with 50 rollout trials, giving 500 rollouts per suite. We report success counts, Wilson 95\% confidence intervals for success rates, and approximate 95\% confidence intervals for absolute improvements over $\pi_{0.5}$.}
\label{tab:statistical_reliability_libero}
\resizebox{\linewidth}{!}{
\begin{tabular}{llccc}
\toprule
Benchmark & Split & $\pi_{0.5}$ & TFP & $\Delta$ vs. $\pi_{0.5}$ \\
\midrule
LIBERO & Spatial
& $494/500 = 98.8\%$ $[97.4,99.4]$
& $498/500 = 99.6\%$ $[98.6,99.9]$
& $+0.8$ pp $[-0.3,+1.9]$ \\
LIBERO & Object
& $490/500 = 98.0\%$ $[96.4,98.9]$
& $495/500 = 99.0\%$ $[97.7,99.6]$
& $+1.0$ pp $[-0.5,+2.5]$ \\
LIBERO & Goal
& $491/500 = 98.2\%$ $[96.6,99.1]$
& $497/500 = 99.4\%$ $[98.3,99.8]$
& $+1.2$ pp $[-0.1,+2.5]$ \\
LIBERO & Long
& $462/500 = 92.4\%$ $[89.7,94.4]$
& $485/500 = 97.0\%$ $[95.1,98.2]$
& $+4.6$ pp $[+1.8,+7.4]$ \\
\textbf{LIBERO} & \textbf{Avg.}
& $\mathbf{1937/2000 = 96.85\%}$ $\mathbf{[96.0,97.5]}$
& $\mathbf{1975/2000 = 98.75\%}$ $\mathbf{[98.2,99.2]}$
& $\mathbf{+1.90}$ pp $\mathbf{[+0.99,+2.81]}$ \\
\bottomrule
\end{tabular}
}
\end{table}

\begin{table}[!htbp]
\centering
\small
\caption{Statistical reliability of real-world Galaxea A1 results. We report Wilson 95\% confidence intervals for success rates over 20 trials per method and task.}
\label{tab:statistical_reliability_real}
\begin{tabular}{llc}
\toprule
Task & Method & Success rate \\
\midrule
Object swap & $\pi_{0.5}$ & $3/20 = 15.0\%$ $[5.2,36.0]$ \\
Object swap & TFP & $15/20 = 75.0\%$ $[53.1,88.8]$ \\
Counting pick-place & $\pi_{0.5}$ & $8/20 = 40.0\%$ $[21.9,61.3]$ \\
Counting pick-place & TFP & $18/20 = 90.0\%$ $[69.9,97.2]$ \\
\bottomrule
\end{tabular}
\end{table}